# Title: A Quantitative Neural Coding Model of Sensory Memory

**Authors:** Peilei Liu[1]*, Ting Wang[1]

**Affiliations:**

[1]College of Computer, National University of Defense Technology, 410073 Changsha, Hunan, China.

*Correspondence to: plliu@nudt.edu.cn

**Abstract**: The coding mechanism of sensory memory on the neuron scale is one of the most important questions in neuroscience. We have put forward a quantitative neural network model, which is self-organized, self-similar, and self-adaptive, just like an ecosystem following Darwin's theory. According to this model, neural coding is a "mult-to-one" mapping from objects to neurons. And the whole cerebrum is a real-time statistical Turing Machine, with powerful representing and learning ability. This model can reconcile some important disputations, such as: temporal coding versus rate-based coding, grandmother cell versus population coding, and decay theory versus interference theory. And it has also provided explanations for some key questions such as memory consolidation, episodic memory, consciousness, and sentiment. Philosophical significance is indicated at last.

**Main Text:** Great strides have been made in neuroscience and cognitive science (*1-8*). Until now however, the coding mechanism of sensory memory on the neuron scale is still unclear. A gap exists between the molecular and whole brain research (*9*). We wish to bridge this gap through a quantitative coding model. Inspired by the "self-organization" idea (*10, 11*), we only make local rules about neuron and synapse based on existing data and theories. Then the hierarchical neural network will emerge automatically. It has features of considerable interest in neuroscience, cognitive science, psychology, and even philosophy. The modeling target is mainly the cerebral cortex. All $c_i$ are constants, which have different meanings in different paragraphs.

**Single neuron model (N)**: N1) $\frac{dq}{dt} = c_1(1-e^{-c_2\sigma})$, $\sigma = m\sum_i p_i a_i$, where q is the ion quantity, m is neuromodulator factor, $p_i$ is the postsynaptic signal, $a_i$ is the activation or sensitivity of membrane channels. The neuron will be fired when $q > c_0$. And the postsynaptic AP (action potential) will induce LTP (long-term potentiation) in this case. Otherwise LTD (long-term depression) will be induced when lacking postsynaptic AP. N1 is inspired by the MP model (*12*) and Hodgkin-Huxley model (*13*). N2) $\frac{dp_i}{dt} = c_3 w_i f_i - c_4 p_i (c_3 \gg c_4)$, $s_p = c_5 p_i f$, where $w_i$ is the synaptic strength, $f_i$ is the presynaptic spike frequency, $s_p$ is stimulus quantity inducing LTP when the neuron is fired, f is the neuron's firing frequency. Namely $p_i$ is determined by two processes: linearly increase of $f_i$ and exponentially decay. N3) $\frac{da_i}{dt} = c_6(c_7 - a_i) > 0 (c_6 \gg 0)$, $s_d = c_8 p_i (c_7 - a_i) = c_9 e^{-c_6 t}$, where $s_d$ is stimulus quantity inducing LTD when the neuron can't be fired. LTD should be caused by stimulus to fatigue channels namely $c_7 - a_i$. Postsynaptic AP can result in complete fatigue of channels namely $a_i = 0$, while EPSP (excitatory postsynaptic potential) will result in partial fatigue namely $0 < a_i < c_7$. Without stimulus however, they will recover to $c_7$ quickly since $c_6 \gg 0$.

N1 and N2 reflect the widely acknowledged spatial and temporal summarization of EPSP. N3 is actually an adaptive mechanism preventing spike frequency becoming too high. This model is supported by experiments and STDP (Spike timing Dependent Plasticity) model (*14*). As in N1, experiments support that inducing LTP needs the coincidence of back-propagating AP and EPSP (*15*). Correspondingly, inducing LTD needs stimulus without postsynaptic AP. According to N1, this require stimulus of low frequency (*16*) or closely after postsynaptic AP (namely stimulus in the refractory period) (*14*). Accordingly to N2 and N3, the time interval between presynaptic and postsynaptic spikes is important. Specifically, the temporal curves of LTP and LTD in STDP should be caused by the EPSP decay and channels recovery respectively. On the other hand however, this neuron model also supports rate-based coding or Hebb coding. For a usual spike sequence, it can be inferred from N2 that $p_i = c_1 f_i (1 - e^{-c_2 t})$, namely $p_i \approx c_1 f_i$ when $t \gg 0$. Therefore $s_p = c_3 p_i f = c_4 f_i f$, and this is actually the Hebb conjecture (*17*). This can explain why inducing LTP needs stimulus of high frequency or collaboration of multiple dendrites (*16*). In conclusion, this model has reconciled the disputations between rate-based coding (Hebb coding) and temporal coding (STDP). Specifically, it is temporal coding for single isolated spike, but rate-based coding for natural spike sequence. We mainly discuss natural spike sequence in this paper, because the isolated spike is usually man-made rather than natural.

For usual spike sequence, $\sigma$ in N1 is approximately constant. Therefore N1 becomes $q = c_1 (1 - e^{-c_2 \sigma}) t$. Namely, the neuron's firing frequency is $f = 1/t = c_3 (1 - e^{-c_2 \sigma})$. Some early models such as MP model also used the form $f = \varphi(\sum_i w_i f_i)$. But the specific form of $\varphi$ here is important for the statistical significances. Suppose that the probability of an object O occurring on condition of attribute $A_i$ is $P(O_i) = P(O|A_i)$, where $A_i$ are independent events. Let $P(\neg O_i) = e^{-c_4 p_i}$, and then $P(O) = P(O_1 + O_2 + \ldots + O_i) = 1 - P(\neg O_1 \neg O_2 \ldots \neg O_i) = (1 - e^{-c_4 \sigma})$, where $\sigma = \sum_i p_i$. Compared with N1, the firing frequency f is proportional to the probability of an object occurring P(O) on condition of attributes. Similarly in N2, $p_i = c_5 f_i (1 - e^{-c_6 t})$ could be viewed as the probabilistic estimate of an attribute occurring $P(A_i)$ on condition of presynaptic spikes history. Without presynaptic spikes however, the reliability of this probabilistic estimate should drop exponentially with time. In essence, N1 and N2 reflect space-time localization which is also a frequently-used principle in the memory management of computer. Namely objects tend to exist in continuous local space and time. In conclusion, neuron is real-time statistical machine. This is also inspired by the statistical physics (*18*) and prediction of John Von Neumann (*19*). Incidentally, the exponential functions in this model can be easily implemented in physics and biology. For example, the radioactive materials decay exponentially.

**Synapse model (S)**: S1) LTP: $\frac{dw}{dt} = c_1 s_p h(c_2 - w) > 0$, $\frac{dr}{dt} = c_3 s_p h(1 - r) > 0$, and $\frac{dw}{dt} = c_4 w \log(r) < 0$ when $s_p = 0$, where w is the synaptic strength, r is the synaptic decay rate, h is hormone factor, $s_p \approx c_5 f_i f$ according to N2. Generally speaking, synapses are alive and self-adaptive like muscles: continuous exercises make them thick and tough, but they will decay passively without exercises. S2) LTD: $\frac{dw_d}{dt} = -c_6 s_d w_d \, (0 < w_d < 1), \frac{dr_d}{dt} = -c_7 s_d r_d \, (0 < r_d < 1),$ and $\frac{dw_d}{dt} = -c_8 \log(r_d)(1-w_d)$ when $s_d = 0$, where $w_d$ is the synaptic devaluation, $r_d$ is the recovery rate of $w_d$. Generally speaking, LTD is the devaluation of LTP, and the actual synaptic strength should be their product $ww_d$. Interestingly, curves in S2 are almost the reversal of S1.

Since LTP is long-term (*16*), stimulus must have changed the synaptic persistence as well as strength. And obviously LTP decays with time. This synapse model is inspired by the BCM model in that synaptic growth rate is influenced by the synaptic strength itself (*20*). From the statistical viewpoint, synaptic strength reflects the confidence of attribute based on stimulus history. Similar to N2, S1 actually reflect the temporal localization principle, which should be the physiological foundation of "recency effect" or "recency bias". Differently however, N2 is a kind of instantaneous memory which could be the foundation of consciousness according to Francis Crick (*1*). We mainly discuss LTP here, because LTD is similar. In addition, LTD is induced by stimulus below the firing threshold. Therefore it should be a regulator for depressing noises other than kernel encoding mechanism (*21*), just like the fatigue of photoreceptors in the retina. Both synaptic increase and decrease are necessary for many neural network models (*6, 7*). According to our model however, the synaptic decrease should be due to passive decay instead of LTD.

**Competition model (C):** C1) $\frac{dp_i}{dt} = -p_d$, $p_d = c_1(1 - e^{-c_2 \sigma})$, $\sigma = \sum_{j=1, j \neq i}^{n} f_j$, where $p_d$ is the quantity of retrograde messengers, n is the number of dendrites connected to the same axonal branch, $p_i$ and $f_i$ are the postsynaptic potential and spike frequency respectively. Specifically, back-propagating AP will release retrograde messengers to the presynaptic and transitorily depress other synapses connecting to the axonal branch (*22*). And their postsynaptic APs will be depressed reversely according to N1. As statistics, this "rich-get-richer" will lead to lateral inhibition known as "softmax" or "winner-take-all" in existing theories (see Fig. 1) (*4, 6*). However, our model doesn't need extra "MAX" layers or lateral inhibitory neurons, which is consistent with physiological evidences. The presynaptic axonal branches themselves actually play similar roles. This convergent competition should be the physiological foundation of "primacy effect" in psychology. And the unique winner is actually the coding neuron of this input, while its revivals can be viewed as background noises. After all, two clocks don't give more accurate time. Form this viewpoint, lateral inhibition is a spatial adaptive mechanism, just like LTD being the temporal adaptive. On the other hand, neurons tend to keep sematic distances apart and prevent replicative coding due to lateral competition. In brief, neurons are professionalized in the division of work. And this will lead to sparse coding (*23*) and the structure of attribute slot defined as following.

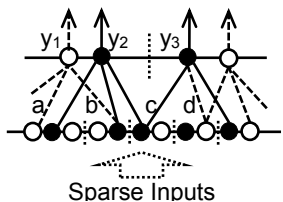
Sparse Inputs

**Fig. 1. Synaptic competition and attribute slot.** Black and white dots mean fired and unfired neurons respectively. Synapse a and d aren't activated for lack of presynaptic spike. But b isn't activated for its failure in the synaptic competition. However, c will be activated somewhat because $y_3$ will be fired by other inputs. As a result, neuron $y_2$ will inhibit $y_1$ known as lateral inhibition, and $y_2$ and $y_3$ will inhibit each other. Lateral inhibition will lead to attribute slots, which are divided by vertical dotted lines here. For example, $y_1$ and $y_2$ are in the same attribute slot, while $y_3$ is in a different one.

**Definition 1:** as in Fig. 1, an attribute slot is a set of neurons: $s=(a_1, a_2, …, a_n)$: if $a_i>0$, then $a_j=0$ for all $j \neq i$, where $a_i$ is the neuron's firing frequency. For example, if every color of a point corresponds to a single detector, all these detectors can compose an attribute slot. Specially, a binary bit is actually a special attribute slot with two values, and it needs two neurons for representing 0 and 1 respectively (see Fig. 2). In some models, 0 is represented by the resting potential. In our opinion however, the resting potential is meaningless because it can't transit neurotransmitters. Therefore the spike is actually unitary other than binary, different from the binary pulse in computer. On the other hand, an attribute with n values can also be represented by the combination of m independent neurons $(\log_b(n) \leq m \leq n)$ other than by n mutually-exclusive neurons. For example, all colors can be mixed of three-primary colors in different proportions. In essence, these two coding manners are "grandmother cell" and population coding respectively (*24*).

Attribute slot has great representative ability. The "XOR problem" once caused the "artificial intelligence winter" (*25*), which was solved by multiple-layered network latter. With attribute slot however, it can be solved through a single-layer network (see Fig. 2). In fact, it can be proved that a single-layer network can represent any logic expression P. As well known, $P = p_1 \vee … p_i \vee … \vee p_n, p_i = a_1 \wedge … a_j \wedge … \wedge a_m$, where $a_j$ is either an atom expression or its negative form. Since $a_j$ and $\neg a_j$ can be represented by attribute slot, $p_i$ can be represented by a single neuron. Therefore P can be represented by a single-layer network. Moreover, inhibitory neurons aren't necessary, different from logic gates in computer. Retrograde massagers actually play similar roles of logic "NOT". Similar to the all-or-none AP, the attribute slot is a kind of digital coding in essence. The digital coding is used widely in computer science for its tolerance of noise. Similarly, the cerebrum can also ignore flaws known as brain completion. The cost is that more neurons are needed.

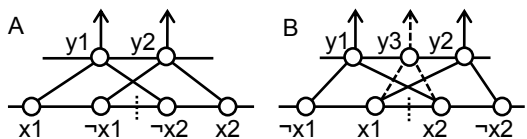

**Fig. 2. Representation of XOR function.** Panel A and B represent "x1 XOR x2" and "x1 OR x2" respectively. Feature each of x1 and ¬x1 corresponds to an independent neuron respectively. In some models however, ¬x1 and ¬x2 are represented by resting potentials.

**Self-organized neural network and circuits.** The initial model is composed of free neurons. Free dendrites move randomly and connect to axonal branches nearby with an initial synaptic strength. Whenever meeting an input, neurons compete and only winners can strengthen their dendritic synapses. Other synapses will decay until broken. Therefore, any input is actually either encoded by a free neuron or merged into the most similar coding neuron. Since single neuron's dendrites are limited, every neuron only encodes a small part of an input. Therefore fine-grain encoding is supported. And the whole input is actually encoded by a hierarchy tree structure, whose root is the coding neuron of this input (see Fig. 3A). Overlapping coding trees compose neural network or cortical column (*26*). Since input is sparse (*23*), for simplicity we suppose that every input is composed of attribute slots ($s_1, s_2, \ldots, s_n$). Then every layer should be composed of attribute slots according to C1. And higher layer contains fewer attribute slots. Therefore every layer is the input of next layer, and feedback fibers can be viewed as common inputs. As results, circuits can be self-organized similar to the feedforward network (see Fig. 3B). Moreover, the neural network is self-similar: coding trees are like large neurons, while columns are large attribute slots. In some degree, a cortical area or the cerebrum itself is a super attribute slot.

According to Francis Crick (*1*), the main cortical structure is determined by genes, while the fine neural coding is determined by postnatal experiences. Consistent with the neural Darwinism theory (*2*), the cortex is like an ecosystem according to our model: inputs are the sunshine; neurons collaborate and compete like plants and animals; and only the fittest can survive. Animals at the top of food chains are actually the memories of environment and era. For instances, lion is the symbol of African grassland, while the tyrannosaurus rex is the memory of the Cretaceous period. Recent experiments support that neural connections change dynamically with external stimulus (*27*). As results, universal free neurons become specialized gradually, just like embryonic stem-cells irreversibly growing into different organs. For example, visual pathways in the developing ferret brain can be rewired to auditory cortex and work well. A neuron's function is actually determined by its dendritic connections. Just as Karl Marx said, "Person is the sum of all social relations". From this viewpoint, people are super neurons of the society or Twitter.

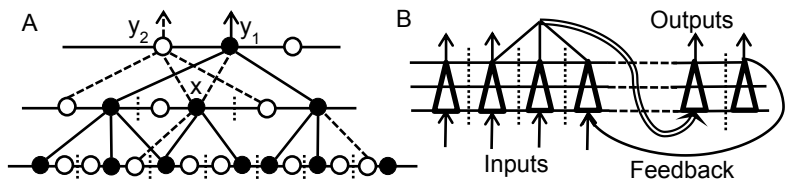

**Fig. 3. Neural network and circuits.** In A, black and white dots mean fired and unfired neurons respectively. Any input will converge to a single neuron such as $y_1$ through a coding tree. Overlapping trees compose a neural network, and common intermediate nodes such as x are shared. Due to lateral inhibition, every layer is composed of attribute slots separated by vertical dotted lines here. In B, the triangle symbol represents coding tree in A. The long feedforward and feedback fibers can be viewed as common inputs. Therefore circuits can be self-organized similar to A.

**Computational significance.** According to our model, neural coding is a "multiple-to-one" mapping from inputs to neurons due to lateral inhibition, which is a well-defined function in essence. An input is encoded only when it has been convergent to the coding neuron through a coding tree. Conversely, whenever a coding neuron is fired, the corresponding input will be retrieved. Obviously, the retrieval process is an encoding process meanwhile. Neurons are computing units as well as storage units. Information is saved in the dendritic synapses, while the firing frequency is actually computing result. From this viewpoint, synapses and spikes are hardware and software of nerve system respectively, and both of them are distributed. Different from the von Neumann architecture (*19*), data and program aren't separated. Representation of information is digital but not binary. In addition, dynamic changes of program at runtime are allowed. From the viewpoint of machine learning, the feedforward encoding is unsupervised learning in general. But supervised learning can also be conducted by facilitating or inhibiting neuron's firing through specific fibers casting. This is important when we need to associate different objects or distinguish similar objects. For instance, we have to associate words and pronunciations with pictures as well as distinguish similar faces. Incidentally, that's why people have "face cells" while monkeys don't (*1*). According to N1 and S1, neuromodulators and hormones are actually meta-mechanisms influencing neuron's firing frequency and learning rate. They could conduct reinforcement learning and selectively reinforcing memories important for survival, similar to the dopamine system in motor learning (*28*). However, the supervised learning and reinforcement learning here are "soft" or statistical rather than absolute. Since fired neurons strengthen their dendritic synapses towards to the input pattern, the lateral competition become convergent and stable gradually. Therefore the neural network tends to converge to the global minimum point, as in the simulated annealing algorithm (*29*). Similarly, synapses become stable gradually with stimuli according to S1. And synapses in lower layers are more stable. As results, iterations in back-propagating algorithm could be avoided when training hierarchical feedforward networks (*7*).

Generally speaking, our model supports "grandmother cell" rather than population coding (*24*). The "grandmother cell" was questioned for following reasons (*1, 4*): first, neurons are too few and too simple for coding infinite complex objects in theory; second, experiments demonstrate that a neuron isn't a template of all stimuli in its receptive field. According to our model however, a neuron encodes more than one object, and not all inputs are stored. Instead, only those occurring frequently or companied with hormone can be saved. And synaptic decay and neurogenesis (*30*) can produce additional memory capacity. Moreover, resources can be saved through sharing common intermediate nodes in the deep hierarchical network (*4, 5*), although retrieval will become slower. And neurons could actually use large trees for encoding complex objects. On the other hand, not all stimuli in a neuron's receptive field are effective. Due to space-time adaptive mechanisms, neurons are more sensitive to features in contrast to backgrounds. Moreover, the coding is also influenced by the vast top-down connections. A possible worry is about the dying of neurons. In our opinion, only those lack of stimulus die. In fact, the influence of accidental death is also limited. As statistical model, lack of partial inputs won't result in fatal disaster to a neuron. And due to the low firing threshold, this neuron model has surprising generalizing ability (*4*). Almost any input could stir up many neurons, and the retrieval is actually determined by lateral competition. Therefore the place of a dead neuron can be taken by the most similar one. In brief, the "grandmother cell" is as robust as population coding. However, population firing confronts the binding problem (*1*), namely how to reconcile distributed information and the unitary consciousness. Brain waves or rhythm was proposed as the binding mechanism. In our opinion however, binding is computing in essence. And neurons themselves should be the workshops binding features together.

**Psychological significance.** S1 actually gives the learning curve and forgetting curve. However, the forgetting curve in experiments isn't consistent with exponential function completely (*31*). In our opinion, recall in the tests and other rehearsals are actually memory consolidation process, which will slow down the decay rate. On the other hand, the lateral inhibition in C1 can explain memories interference (*32*). Since it influences encoding and retrieval meanwhile, both proactive interference and retroactive interference can be explained. Therefore our model has actually reconciled the decay theory and interference theory (*33*). In essence, forgetting is retrieval failure due to either lateral inhibition or disappearance of "memory trace". Some researchers hold that forgetting should result from random changes of synapses due to noise (*34*), because relearning is easier than learning new, namely the implicit memory in psychology. According to our model, implicit memory are stored in lower layers, where neurons are common nodes shared by different coding trees. Therefore their synapses are usually tougher and more likely to remain. And forgetting should mainly be due to synaptic decay near the root of coding trees. Repetitive stimulus (rehearsal) and hormones can lead to memory consolidation. For example, we can remember a strange phone number through quick repetition. And dream should be one of such rehearsals (*35*). Specifically, neurons will become sensitive and firing spontaneously when without external inputs in sleep. Due to lateral inhibition, only strong circuits can be fired. Therefore in essence, dream is the self-reinforcing of strong memories as well as the fantasy divorced from reality since lacking inputs from nature. Influential events are often companied with hormones changing, which actually represent their importance for the survival (*36*). Since an event seldom repeats, it is hard for the neocortex to encode. However, it can be stored in hippocampus and limbic structure which are sensitive to hormones. These memories are usually known as episodic memory or flash memory, which is an important part of working memory in our opinion. They should be able to influence the sematic memory in cortex (*37*), just like computer memory writing and reading disk. From this viewpoint, sematic memory and rationality serve the hormone system, or instinct dominates rationality.

Consciousness is one of the most important and mystical topics (*1, 3, 8*), which has different meanings in different disciplines. Here we try to discuss the neural mechanism correlative with consciousness, namely what happens when you think of an object. According to our model, conscious of something means the corresponding coding neuron's firing. Paying attention to an object means specific fibers and neuromodulators casting to the coding neuron, while sentiment should be pervasive hormones. And different areas should be sensitive to different hormones, as the Lövheim cube of emotion (*38*). Our consciousness is often believed to be autonomous and free. For example, a girl can make various sentences she has never heard of. In our opinion, the consciousness seems autonomous because of the continuous complex inputs, huge memory volume, and some randomness. Specifically, the cerebrum is a "Turing Machine" in essence, whose output is determined by current inputs and internal states. Current inputs come from nature as well as our body, while internal states include memory and instinct system. In fact, the instinct or sentiment can also be viewed as ancient memories encoded in genes. As Crick said, we believe our minds are free, because we often know the decision itself but unaware of the decision-making process (*1*). In other words, the cerebrum is approximately a black box for us. Open the mainframe however, you will find a not-so-intelligent machine composed of transistors and plastic connections, although it is indeed more robust. It is surprising and interesting that intelligence could come from random processes.

The philosophical significance hasn't escaped our notice: the consciousness is the second nature, while the so-called "nature" is actually our subjective cognition of the "real" nature. We can cognize the nature only in probability and statistics, but never precisely or completely. In brief, the materialism and idealism can be reconciled.

**Acknowledgments:** This work was supported by the National Natural Science Foundation of China (Grant No. 61170156 and 60933005).